\documentclass[conference]{IEEEtran}
\IEEEoverridecommandlockouts
\usepackage{url}
\usepackage{cite}
\usepackage{amsmath,amssymb,amsfonts}
\usepackage{graphicx}
\usepackage{textcomp}
\usepackage{xcolor}
\usepackage{multirow}
\usepackage[ruled,vlined]{algorithm2e}
\begin{document}

\title{Robust Sentiment Analysis for Low Resource languages Using Data Augmentation Approaches:\\ A Case Study in Marathi\\
}

\makeatletter
\newcommand{\linebreakand}{%
  \end{@IEEEauthorhalign}
  \hfill\mbox{}\par
  \mbox{}\hfill\begin{@IEEEauthorhalign}
}
\makeatother

\author{
  \IEEEauthorblockN{Aabha Pingle}
  \IEEEauthorblockA{
  % \textit{Department of Computer Engineering} \\
    \textit{\footnotesize SCTR's Pune Institute of Computer Technology}\\
    \textit{\footnotesize L3Cube Pune}\\
    % Pune, India \\
    aabhapingle@gmail.com}
  \and
  \IEEEauthorblockN{Aditya Vyawahare}
  \IEEEauthorblockA{
  % \textit{Department of Computer Engineering} \\
    \textit{\footnotesize SCTR's Pune Institute of Computer Technology}\\
    \textit{\footnotesize L3Cube Pune}\\
    % Pune, India \\
    aditya.vyawahare07@gmail.com}
  \and
  \IEEEauthorblockN{Isha Joshi}
  \IEEEauthorblockA{
  % \textit{Department of Computer Engineering} \\
    \textit{\footnotesize SCTR's Pune Institute of Computer Technology}\\
    \textit{\footnotesize L3Cube Pune}\\
    % Pune, India \\
     joshiishaa@gmail.com}
  \linebreakand % <------------- \and with a line-break
  \hspace{-0.6cm}
  \IEEEauthorblockN{Rahul Tangsali}
  \IEEEauthorblockA{
  % \textit{Department of Computer Engineering} \\
    \textit{ \footnotesize SCTR's Pune Institute of Computer Technology}\\
    \textit{\footnotesize L3Cube Pune}\\
     % Pune, India \\
     rahuul2001@gmail.com}
    \and
    \IEEEauthorblockN{Geetanjali Kale}
  \IEEEauthorblockA{
  % \textit{Department of Computer Engineering} \\
    \textit{ \footnotesize SCTR's Pune Institute of Computer Technology}\\
    % Pune, India \\
    gvkale@pict.edu}
   
  \and
   \IEEEauthorblockN{Raviraj Joshi}
  \IEEEauthorblockA{
  % \textit{Department of Computer Science and Engg} \\
    \textit{\footnotesize Indian Institue of Technology Madras}\\
    \textit{\footnotesize L3Cube Pune}\\
    % Chennai, India \\
    ravirajoshi@gmail.com}
     \\
  
}

\maketitle

\begin{abstract}
Sentiment analysis plays a crucial role in understanding the sentiment expressed in text data. While sentiment analysis research has been extensively conducted in English and other Western languages, there exists a significant gap in research efforts for sentiment analysis in low-resource languages. Limited resources, including datasets and NLP research, hinder the progress in this area. In this work, we present an exhaustive study of data augmentation approaches for the low-resource Indic language Marathi. Although domain-specific datasets for sentiment analysis in Marathi exist, they often fall short when applied to generalized and variable-length inputs. To address this challenge, this research paper proposes four data augmentation techniques for sentiment analysis in Marathi. The paper focuses on augmenting existing datasets to compensate for the lack of sufficient resources. The primary objective is to enhance sentiment analysis model performance in both in-domain and cross-domain scenarios by leveraging data augmentation strategies. The data augmentation approaches proposed showed a significant performance improvement for cross-domain accuracies. The augmentation methods include paraphrasing, back-translation; BERT-based random token replacement, named entity replacement, and pseudo-label generation; GPT-based text and label generation. Furthermore, these techniques can be extended to other low-resource languages and for general text classification tasks.
\end{abstract}


\section{Introduction}
% need for sentiment analysis for low-resource languages
Low-resource languages lack the tools necessary to perform computational activities like sentiment analysis, such as annotated data and language models \cite{joshi2022l3cubemahanlp}. On social media, many low-resource language communities are becoming more active online and debating social and political issues. Social and political topics are frequently addressed online in low-resource language groups on sites like Twitter, Facebook, etc. Low-resource languages are frequently related to developing economies, where companies may desire to comprehend consumer attitudes and preferences, particularly with regard to the Indian subcontinent. Therefore, there exists a need for exhaustive research in sentiment analysis for low-resource languages.
% importance of data augmentation in sentiment analysis
\\ \indent Data augmentation is a machine learning technique that creates new synthetic data from existing data to boost the quantity and variety of a dataset. Data augmentation, which provides additional instances for underrepresented sentiment labels, can be used to resolve unbalanced sentiment analysis datasets. By offering fresh patterns and variations for the model to learn from, this strategy helps to reduce overfitting and increases model generalization. Furthermore, by generating new data with variations in spelling, word order, and other natural language features, data augmentation can improve model robustness.
% marathi as a language
\\ \indent Marathi is an Indo-Aryan language mainly spoken in the Indian state of Maharashtra and some neighboring states such as Goa, Karnataka, and Gujarat. With 83 million speakers in 2011 and exponentially increasing ever since, it is one of the 22 scheduled languages of India, ranking 13th globally and 3rd in India after Hindi and Bengali. Despite its significant number of speakers worldwide, Marathi is considered a low-resource language. One of the most significant issues is a lack of appropriate datasets and morphological analyzers in Marathi, both of which are required for constructing accurate and effective NLP models. Furthermore, Marathi is frequently domain-dependent in NLP tasks, making it difficult to generalise models across domains. This makes developing high-quality NLP tools and technologies for Marathi challenging, restricting its use and uptake in a variety of applications.

\section{Related Work}

The goal of sentiment analysis, often known as opinion mining, is to automatically recognize and extract subjective information from text data \cite{Dang_2020}. Using this knowledge, one may learn more about how individuals feel and behave toward certain issues. Numerous applications, including social media monitoring, customer feedback analysis, and market research, can benefit from sentiment analysis \cite{DRUS2019707}. Sentiment analysis has become a crucial technique for determining how the public feels about various goods, services, and events as a result of the increased accessibility of text data on the internet.
\\
\indent Text preprocessing, feature extraction, and sentiment classification are just a few of the NLP activities that go into sentiment analysis. To clean and prepare the text data for subsequent analysis, text pre-processing comprises operations like tokenization, stemming, and stop-word removal \cite{HADDI201326}. As part of feature extraction, text data is transformed into a numerical representation that may be fed into a machine learning model. Finding the text's sentiment polarity—whether it is positive, negative, or neutral—is the problem of sentiment classification. Usual performance criteria for sentiment categorization models include accuracy, precision, recall, and F1-score.
\\
\indent Convolutional Neural Networks (CNN) \cite{7363395} and Recurrent Neural Networks (RNN), two deep learning algorithms, have significantly advanced the area of sentiment analysis in recent years . These methods have demonstrated efficacy in enhancing the functionality of sentiment analysis models across a range of languages and topics. But there are still a lot of difficulties to be solved, such as handling subjectivity, irony, and denial in text data. Additionally, Marathi sentiment analysis is still a study area that has to be explored. Despite these difficulties, sentiment analysis is anticipated to keep playing a significant role in the big data age, offering insightful information about people's attitudes and views.\\

In natural language processing, data augmentation is commonly used to expand the quantity or diversity of a dataset. It is usually done to tackle problems involving a dearth of textual data. By artificially generating data and different versions of the dataset, the model can improve its generalization and robustness, resulting in better performance on unseen data. Data augmentation techniques can be used in sentiment analysis to add variations to a dataset, while ensuring that the sentiment of the sentences is retained.\\

\indent Data augmentation has been widely used in Computer Vision to expand the dataset and improve model performance. However, data augmentation in NLP is challenging as it requires a careful balance of expanding the dataset and preserving the language's grammar, syntax, and lexical semantics. Hence, the data augmentation technique used should be selected such that it does not affect the model performance and does not introduce unwanted bias or noise. 

\subsection{Lexical Substitution}\indent \cite{wei-zou-2019-eda} introduced lexical substitution that involves substituting randomly chosen words with their synonym. Additionally, the paper proposes several other easy data augmentation (EDA) techniques like random insertion, random swap, and random deletion which involve randomly selecting words and performing a specific operation on them. These techniques have proven to provide a boost in performance for NLP tasks, especially for smaller datasets where overfitting is a common concern. 

\subsection{Query Expansion}\indent Query Expansion can also be used to generate data for data augmentation. Azad and Deepak (2019) \cite{azad2019query} describe how to generate semantically similar query terms from an initial query so as to retrieve more relevant documents. 

\subsection{Mixup}\indent The data augmentation technique called mixup presented by Zhang et al. (2017) \cite{zhang2017mixup} originally introduced for image recognition tasks involves linearly interpolating two samples of data and their corresponding labels to predict the labels. Given that word embeddings in NLP are similar to feature vectors in computer vision, mixup can be adapted for use in NLP tasks to augment textual data. 

\subsection{Back-Translation}\indent The back-translation method, first introduced in the field of machine translation by  Sennrich et al. (2016) \cite{sennrich-etal-2016-improving} and further developed by  Sugiyama and Yoshinaga (2019)\cite{sugiyama-yoshinaga-2019-data}, involves translating one language to a different language or dialect and back again. Back-translation has proven to improve machine translation for low-resource languages. Furthermore, back-translation can be potentially applied to other NLP tasks like sentiment analysis by generating sentences that capture different dialects and languages, helping the model to generalize better. 

\subsection{Data Denoising}\indent Data denoising is a data augmentation method that involves intentionally adding noise to the input data. The noise can be introduced by misspelling words, adding typos, or grammatical errors such that model becomes less susceptible to common errors. Ng et al. (2020)\cite{ng-etal-2020-ssmba} presents a similar augmentation technique called SSMBA (Self-Supervised Manifold Based Data Augmentation) that uses a pair of corruption and reconstruction functions to generate synthetic training examples. 

\subsection{Text Paraphrasing and Summarization}\indent Text paraphrasing and text summarization are additional data augmentation techniques that can be used to expand the dataset. Text paraphrasing involves changing some words while retaining the meaning of the sentence, whereas text summarization involves condensing the corpus to a shorter and clearer form. By using these methods, data can be expanded while allowing the model to learn different vocabulary from varying word usage and text lengths. 

\section{Datasets used}
In our research, we conducted experiments using two datasets, namely GoEmotions \cite{demszky-etal-2020-goemotions} and L3CubeMahaSent \cite{DBLP:journals/corr/abs-2103-11408}, to evaluate the effectiveness of our proposed augmentation techniques. To ensure a comprehensive analysis, we utilized the validation and test sets from both datasets. This allowed us to perform a thorough evaluation of our methods in both cross-domain and in-domain scenarios.
\subsection{MahaSent}
L3CubeMahaSent is a sentiment analysis dataset in Marathi released by L3Cube. The dataset consists of Marathi tweets and was compiled by scraping tweets from various Marathi individuals. The tweets are about current events, and the dataset includes several tweets from politicians and activists. It comprises approximately 16,000 tweets that have been classified as negative, neutral or positive. The training dataset consists of 12,114 tweets that can be used to train sentiment analysis models. Furthermore, the test and validation datasets contain 2,250 and 1,500 tweets, respectively \cite{DBLP:journals/corr/abs-2103-11408}.
\subsection{GoEmotions}
The GoEmotions dataset is a  fine-grained emotions dataset originally compiled in English. It consists of 58,000 Reddit comments in the English language, covering a wide range of topics and discussions. The dataset includes 28 distinct emotion labels, including neutral, resulting in a total of 27 specific emotions. For the purpose of our research work, the GoEmotions dataset has been translated to Marathi. Additionally, the 28 emotion labels were mapped to a three-point scale, namely positive, negative, and neutral \cite{demszky-etal-2020-goemotions}.
\section{System description}
We first preprocessed the MahaSent and GoEmotions datasets by removing hashtags, punctuations, links and non-devanagri content from every sentence. We then employed the marathi-bert-v2 model as the baseline model for our experiments. This pre-trained BERT model was fine-tuned on both the preprocessed datasets to create two distinct models. To establish the performance baseline, we evaluated the accuracy scores of both models on the test and validation sets from both the GoEmotions and MahaSent datasets. These baseline scores provided a reference point for assessing the effectiveness of the data augmentation techniques employed in our study. By comparing the performance of augmented models with the baseline scores, we were able to measure the impact and improvement achieved through the data augmentation techniques in our research.

Initially, we used two existing data augmentation techniques, namely back-translation \cite{sennrich-etal-2016-improving} and paraphrasing for experimentation. However, we were able to observe an improvement in performance (with respect to accuracy scores) using the proposed augmentation methods. The following sections describe all the approaches used for data augmentation.
\subsection{Back-translation}
This approach leverages machine translation models to augment text data. This technique involves translating sentences from the target language to a different language using a machine translation model and then translating them back to the original language. By performing this translation, the original sentences are transformed into new variations while preserving the original meaning. We performed back-translation on Marathi text using English as the intermediate language.\\

\begin{algorithm}[]
\SetAlgoLined
\KwIn{input\_sentence: A string representing the input sentence to be performed back-translation with}
\KwOut{backTranslatedSentence: A string representing the sentence obtained after back-translation}
englishSentence $\leftarrow$ translateToEnglish(input\_sentence)\;
backTranslatedSentence $\leftarrow$ translateToMarathi(englishSentence)\;
% modified\_sentence $\leftarrow$ Join(word\_list)\;
\Return backTranslatedSentence\;
\caption{Back-Translation}
\end{algorithm}
\subsection{Paraphrasing}
Paraphrasing is a data augmentation technique that involves rephrasing sentences while retaining their original meaning. Using this method, we aim to generate new variations of existing sentences. Paraphrasing enriches the dataset by introducing different sentence structures, synonyms, and expressions. We used the ai4bharat/MultiIndicParaphraseGeneration \footnote{https://huggingface.co/ai4bharat/MultiIndicParaphraseGeneration} model to perform the sentence wise paraphrasing of the Marathi text in our datasets.\\

\begin{algorithm}[]
\SetAlgoLined
\KwIn{input\_sentence: A string representing the input sentence to be performed back-translation with}
\KwOut{paraphrasedSentence: A string representing the sentence obtained after paraphrasing the input sentence using the MultiIndicParaphrasingGeneration model.}
paraphrasedSentence $\leftarrow$ paraphrase(input\_sentence)\;
\Return paraphrasedSentence\;
\caption{Paraphrasing}
\end{algorithm}
\subsection{Random Masking}
\subsubsection{Sequential}
In sequential random masking using BERT, 40\% of the tokens in the sentence were randomly chosen. The first randomly selected word was masked and then the marathi-bert-v2 model \footnote{https://huggingface.co/l3cube-pune/marathi-bert-v2} was used to predict the masked token. The token was then replaced in the original sentence with the predicted token. Once the replacement is done, the same procedure is repeated for all other selected tokens.\\

\begin{algorithm}[]
\SetAlgoLined
\KwIn{input\_sentence: A string representing the input sentence to be masked}
\KwOut{modified\_sentence: A string representing the input sentence after sequential random masking}
\hspace{0mm}word\_list $\leftarrow$ Split(input\_sentence)\;
num\_masked\_words $\leftarrow$ Round(0.4 $\times$ Length(word\_list))\;
\For{i $\leftarrow$ 1 to num\_masked\_words}{
    \hspace{0mm}index $\leftarrow$ RandomlySelectIndex(word\_list)\;
    word $\leftarrow$ word\_list[index]\;
    masked\_word $\leftarrow$ MaskWord(word)\;
    predicted\_word $\leftarrow$ BERT.predict(masked\_word)\;
    word\_list[index] $\leftarrow$ predicted\_word\;
}
modified\_sentence $\leftarrow$ Join(word\_list)\;
\hspace{0cm}\Return modified\_sentence\;  
\caption{Sequential Random Masking using BERT}
\end{algorithm}

\subsubsection{Parallel}
Similarly, for parallel random masking, 40\% of the words were randomly selected for masking. Each selected word was masked separately and tokens were predicted using the marathi-bert-v2 model. The tokens are then simultaneously replaced by all the predicted tokens obtained in parallel \\

\begin{algorithm}[]
\SetAlgoLined
\KwIn{input\_sentence: A string representing the input sentence to be masked}
\KwOut{modified\_sentence: A string representing the input sentence after parallel random masking}
word\_list $\leftarrow$ Split(input\_sentence)\;
num\_masked\_words $\leftarrow$ Round(0.4 $\times$ Length(word\_list))\;
predicted\_words $\leftarrow$ Empty list\;
\For{i $\leftarrow$ 1 to num\_masked\_words}{
    index $\leftarrow$ RandomlySelectIndex(word\_list);
    word $\leftarrow$ word\_list[index]\;
    masked\_word $\leftarrow$ MaskWord(word)\;
    predicted\_word $\leftarrow$ BERT.predict(masked\_word)\;
    predicted\_words.Append(predicted\_word)\;
}
modified\_sentence $\leftarrow$ ReplaceMaskedWords(word\_list, predicted\_words)\;
\Return modified\_sentence\;
\caption{Parallel Random Masking using BERT}
\end{algorithm}
\subsection{Named Entity Masking}
\subsubsection{Sequential}
In sequential NER masking using BERT, all of the tokens in the sentence were first tagged using the MahaNER-BERT model \footnote{https://huggingface.co/l3cube-pune/marathi-ner}. The first named entity was masked and then the marathi-bert-v2 model was used to predict the masked token. The token was then replaced in the original sentence with the predicted token. Once the replacement is done, the same procedure is repeated for all named entities.

\begin{algorithm}[]
\SetAlgoLined
\KwIn{Sentence $S$}
\KwOut{Modified sentence $S'$ with replaced named entities}
NamedEntities $\leftarrow$ MahaNLP.extractNamedEntities($S$)\;
\For{NamedEntity $NE$ in NamedEntities}{
    \If{NE.category $\neq$ 'Other'}{
        MaskedNE $\leftarrow$ MaskToken(NE)\;
        PredictedWord $\leftarrow$ BERT.predict(MaskedNE)\;
        $S' \leftarrow$ ReplaceToken($S'$, MaskedNE, PredictedWord)\;
    }
}
\Return{$S'$}\;
\caption{Sequential Named Entity Masking}
\label{algo:ner_algorithm}
\end{algorithm}

\subsubsection{Parallel}
Similarly, for parallel NER masking, all of the tokens were tagged using the MahaNER-BERT model. Each named entity was masked separately and tokens were predicted using the marathi-bert-v2 model. The tokens are then simultaneously replaced by all the predicted tokens obtained in parallel.

\begin{algorithm}[]
\SetAlgoLined
\KwIn{input\_sentence: A string representing the input sentence to be masked}
\KwOut{modified\_sentence: A string representing the input sentence after parallel named entity masking}
NamedEntities $\leftarrow$ MahaNLP.extractNamedEntities($S$)\;
% ner\_tokens $\leftarrow$ Round(0.4 $\times$ Length(word\_list))\;
index\_list $\leftarrow$ Empty list\;
predicted\_words $\leftarrow$ Empty list\;
\For{NamedEntity $NE$ in NamedEntities}{
    \If{NE.category $\neq$ 'Other'}{
        index $\leftarrow$ getIndex(NE)\;
        index\_list.Append(index)\;
        masked\_word $\leftarrow$ MaskWord(NE)\;
        predicted\_word $\leftarrow$ BERT.predict(masked\_word)\;
        predicted\_words.Append(predicted\_word)\;
    }
    
}
modified\_sentence $\leftarrow$ ReplaceMaskedWordsWithIndex(predicted\_words, index\_list)\;
\Return modified\_sentence\;
\caption{Parallel Named Entity Masking}
\end{algorithm}

\subsection{BERT-based approach}
In this pseudo-labeling approach, we use the marathi-bert-v2 model, a BERT-based model for generating labels on a given set of data. This model was first fine-tuned on the MahaSent dataset for the sentiment analysis task. After that, labels were generated on the GoEmotions dataset. The generated labels were one of the sentiments, i.e. positive, negative, or neutral. We use the previous checkpoint of the marathi-bert-v2 to further fine-tune it on this newly generated data. The accuracy of this model was checked on the test and validation sets of both the Goemotions as well as the MahaSent dataset. This approach focuses on the cross-domain analysis of the model. 

\begin{algorithm}[]
\SetAlgoLined
\KwIn{input\_sentence: A string representing the input sentence to be masked}
\KwOut{label: A target label representing the sentiment of the input sentence. The output will be either one of these integers 0, 1, and 2 representing negative, positive, and neutral sentiments respectively. 
}
preprocessed\_sentence $\leftarrow$ preprocess(input\_sentence)\;
tokenised\_sentence $\leftarrow$ tokenise(input\_sentence)\;
label $\leftarrow$ BERT.predict(tokenised\_sentence)\;
sample $\leftarrow$ input\_sentence, label\;
\Return sample;

\caption{BERT-based label generation approach}
\end{algorithm}
 
\subsection{GPT-based approach 1}
\hspace{1.27cm} In this particular approach, we employ the usage of the 'l3cube-pune/marathi-gpt' model, which is based on GPT (Generative Pre-trained Transformer), to generate labels for a specific dataset. This model, known as GPT2, has undergone pre-training on L3Cube-MahaCorpus \cite{joshi2022l3cubemahacorpus}, a vast collection of Marathi language data. Subsequently, we employed the trained GPT model to predict sentences on a distinct domain dataset, and we evaluated the accuracy of these predictions on both the test and validation sets of both the Goemotions and MahaSent datasets. The primary focus of this approach lies in conducting a comprehensive analysis of the model's performance in both cross-domain and in-domain scenarios. By evaluating its ability to generate accurate labels on data from different domains, we can gauge the model's adaptability and effectiveness in handling various types of inputs. This analysis serves as a crucial step in understanding the strengths and limitations of the GPT-based model and its applicability in different contexts. \\

\begin{algorithm}[]
\SetAlgoLined
\KwIn{input\_sentence: A string representing the input sentence}
\KwOut{label: A target label representing the sentiment of the input sentence. The output will be either one of these integers 0, 1, and 2 representing negative, positive, and neutral sentiments respectively. 
}
preprocessed\_sentence $\leftarrow$ preprocess(input\_sentence)\;
tokenised\_sentence $\leftarrow$ tokenise(input\_sentence)\;
label $\leftarrow$ GPT.predict(tokenised\_sentence)\;
sample $\leftarrow$ input\_sentence, label\;
\Return sample;
\caption{GPT-based Approach 1}
\end{algorithm}

\subsection{GPT-based approach 2}

In this specific approach, we utilize the 'l3cube-pune/marathi-gpt' model, which is built upon the powerful GPT architecture. Rather than generating complete sentences from scratch, the GPT model was trained to intelligently complete partial sentences without altering the original sentiment conveyed by the sentence. To facilitate this, we begin by creating a dataset that includes halved sentences extracted from the original sentences. These partial sentences serve as prompts for the GPT model to generate meaningful and sentiment-consistent completions. These newly generated sentences were combined with the original dataset, resulting in an augmented dataset that encompassed a broader range of labeled examples. The augmented dataset, comprising both the original and the GPT-generated sentences, was used to train the model. Subsequently, we evaluated the accuracy of the model's predictions on both the test and validation sets of the Goemotions and MahaSent datasets. This evaluation allowed us to assess the performance of the model in generating accurate and sentiment-aligned labels on various test examples, providing valuable insights into its effectiveness and generalization capabilities.\\

\begin{algorithm}[]
\SetAlgoLined
\KwIn{input\_sentence: A label and halved string that represents a partial version of the original input sentence.}
\KwOut{string: A final completed sentence that retains the sentiment of the original input while offering additional context or completion.
}
preprocessed\_sentence $\leftarrow$ preprocess(input\_sentence)\;
tokenised\_sentence $\leftarrow$ tokenise(input\_sentence)\;
label $\leftarrow$ GPT.predict(tokenised\_sentence)\;
sample $\leftarrow$ input\_sentence, label\;
\Return sample;
\caption{GPT-based Approach 2}
\end{algorithm}

\begin{table*}[]
\caption{Accuracy Scores for Models Trained Using Data Augmentation Techniques}
\resizebox{\textwidth}{!}{
\begin{tabular}{|l|c|c|c|c|c|c|c|c|}
\hline
\multirow{2}{*}{\begin{tabular}[c]{@{}c@{}} Augmentation Technique Used\end{tabular}} &
  \multicolumn{4}{c|}{Model Fine-Tuned on MahaSent} &
  \multicolumn{4}{c|}{Model Fine-Tuned on GoEmotions} \\ \cline{2-9} 
 &
  \multicolumn{2}{c|}{GoEmotions} &
  \multicolumn{2}{c|}{MahaSent} &
  \multicolumn{2}{c|}{GoEmotions} &
  \multicolumn{2}{c|}{MahaSent} \\ \cline{2-9} 
 &
  \begin{tabular}[c]{@{}c@{}}Validation\\ Set\end{tabular} &
  \begin{tabular}[c]{@{}c@{}}Test\\ Set\end{tabular} &
  \begin{tabular}[c]{@{}c@{}}Validation\\ Set\end{tabular} &
  \begin{tabular}[c]{@{}c@{}}Test\\ Set\end{tabular} &
  \begin{tabular}[c]{@{}c@{}}Validation\\ Set\end{tabular} &
  \begin{tabular}[c]{@{}c@{}}Test\\ Set\end{tabular} &
  \begin{tabular}[c]{@{}c@{}}Validation\\ Set\end{tabular} &
  \begin{tabular}[c]{@{}c@{}}Test\\ Set\end{tabular} \\ \hline
Base Model &
  - &
  \textbf{0.5236} &
  \textbf{0.8587} &
  \textbf{0.8367} &
  \textbf{0.6297} &
  \textbf{0.6263} &
  - &
  \textbf{0.6882} \\ \hline
Back-translation &
  - &
  0.5266 &
  0.8620 &
  0.8358 &
  0.6301 &
  0.6269 &
  - &
  0.7126 \\ \hline
Paraphrasing &
  - &
  0.4962 &
  0.8573 &
  0.8403 &
  0.6352 &
  0.6253 &
  - &
  0.7210 \\ \hline
Named Entity Masking Using BERT (Sequential) &
  - &
  \textbf{0.5372} &
  0.8593 &
  0.8359 &
  0.6359 &
  0.6313 &
  - &
  \textbf{0.7362} \\ \hline
Named Entity Masking Using BERT (Parallel) &
  - &
  0.5275 &
  0.8673 &
  0.8367 &
  0.6344 &
  0.6258 &
  - &
  0.6993 \\ \hline
Random Masking Using BERT (Sequential) &
  - &
  0.5311 &
  0.8580 &
  \textbf{0.8430} &
  0.6271 &
  \textbf{0.6320} &
  - &
  0.7062 \\ \hline
Random Masking Using BERT (Parallel) &
  - &
  0.5364 &
  0.8440 &
  0.8420 &
  0.6310 &
  0.6302 &
  - &
  0.7160 \\ \hline
\end{tabular}
}
\vspace{-4pt}
\label{tab:my-table}
\end{table*}

\begin{table*}[]
\caption{Accuracy Scores for Models Trained using the GPT-based method for label generation}
\resizebox{\textwidth}{!}{%
\begin{tabular}{|cccccccc|}
\hline
\multicolumn{8}{|c|}{GPT based method (label generation)} \\ \hline
\multicolumn{4}{|c|}{In-domain Analysis} &
  \multicolumn{4}{c|}{Cross-domain Analysis} \\ \hline

\multicolumn{2}{|c|}{GoEmotions dataset} &
  \multicolumn{2}{c|}{MahaSent} &
  \multicolumn{2}{c|}{GoEmotions dataset} &
  \multicolumn{2}{c|}{MahaSent} \\ \hline
\multicolumn{1}{|c|}{Validation Set} &
  \multicolumn{1}{c|}{Test Set} &
  \multicolumn{1}{c|}{Validation Set} &
  \multicolumn{1}{c|}{Test Set} &
  \multicolumn{1}{c|}{Validation Set} &
  \multicolumn{1}{c|}{Test Set} &
  \multicolumn{1}{c|}{Validation Set} &
  Test Set \\ \hline

\multicolumn{1}{|c|}{0.5061} &
  \multicolumn{1}{c|}{0.5068} &
  \multicolumn{1}{c|}{0.8633} &
  \multicolumn{1}{c|}{0.8435} &
  \multicolumn{1}{c|}{0.5227} &
  \multicolumn{1}{c|}{0.5249} &
  \multicolumn{1}{c|}{0.8540} &
  0.8400 \\ \hline
\end{tabular}%
}
\vspace{-4pt}
\label{tab:my-table}
\end{table*}

\begin{table*}[]
\caption{Accuracy Scores for Models Trained using the GPT-based Method for Text Generation}
\begin{center}
\resizebox{0.5\textwidth}{!}{%
\centering
\begin{tabular}{|c|c|c|c|}
\hline
\multicolumn{4}{|c|}{GPT-based Method for Text Generation}                              \\ \hline
\multicolumn{2}{|c|}{GoEmotions dataset} & \multicolumn{2}{c|}{MahaSent dataset} \\ \hline
\multicolumn{1}{|c|}{Validation Set} &
  \multicolumn{1}{c|}{Test Set} &
  \multicolumn{1}{c|}{Validation Set} &
  Test Set \\ \hline
\multicolumn{1}{|c|}{0.5121} &
  \multicolumn{1}{c|}{0.5106} &
  \multicolumn{1}{c|}{0.8500} &
  0.8351 \\ \hline
\end{tabular}
}
\end{center}
\vspace{-6pt}
\label{tab:my-table}
\end{table*}

\begin{table*}[]

\centering
\caption{Confusion Matrices for Random Parallel Masking (GoEmotions model)}
\begin{tabular}{|l|l|l|}

\hline
                                     & GoEmotions & MahaSent                       \\ \hline
\multicolumn{1}{|c|}{Validation Set} &  {\includegraphics[scale=0.5]{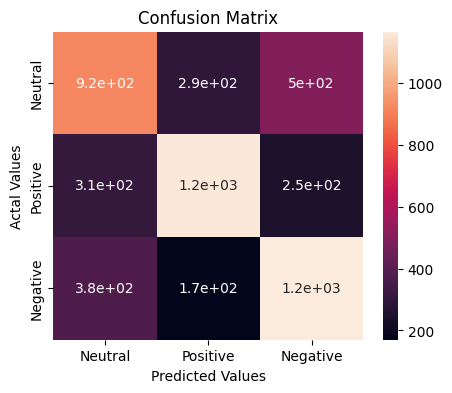}} & {\includegraphics[scale=0.5]{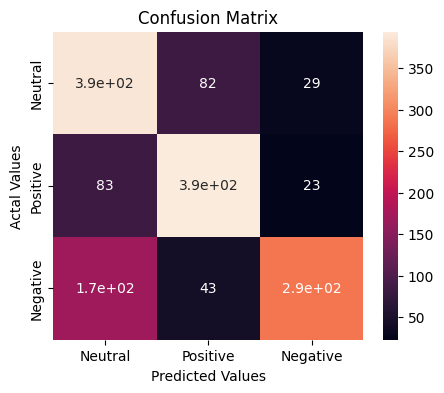}} \\ \hline
Test Set                             &     {\includegraphics[scale=0.5]{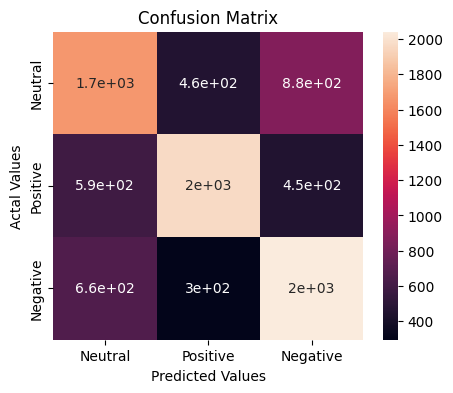}}       &         {\includegraphics[scale=0.5]{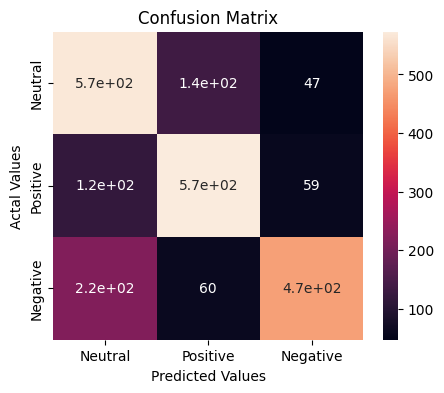}}                       \\ \hline
\end{tabular}

\label{tab:my-table}
\end{table*}

\begin{table*}[]
\caption{Confusion Matrices for Random Sequential Masking (GoEmotions model)}
\centering
\begin{tabular}{|l|l|l|}

\hline
                                     & GoEmotions & MahaSent                       \\ \hline
\multicolumn{1}{|c|}{Validation Set} &  {\includegraphics[scale=0.5]{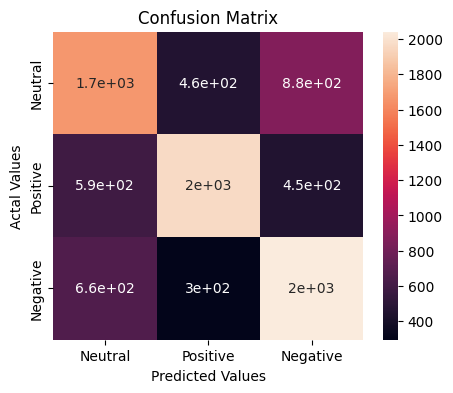}} & {\includegraphics[scale=0.5]{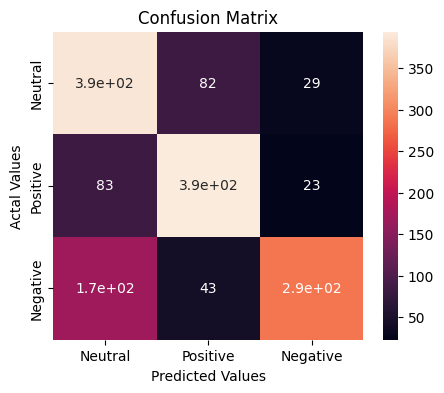}}  \\ \hline
Test Set                             &     {\includegraphics[scale=0.5]{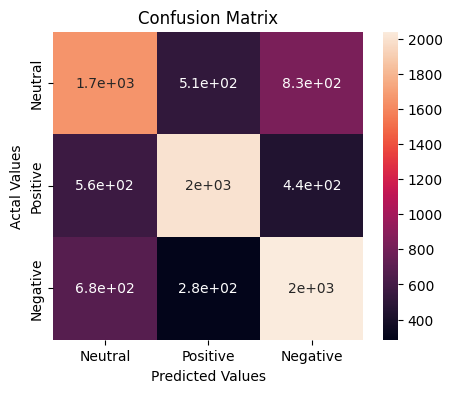}}       &         {\includegraphics[scale=0.5]{go_emotions_sequential_m_val_cm.png}}                       \\ \hline
\end{tabular}

\label{tab:my-table}
\end{table*}

\begin{table*}[]
\caption{Confusion Matrices for Random Parallel Masking (MahaSent Model)}
\centering
\begin{tabular}{|l|l|l|}

\hline
                                     & GoEmotions & MahaSent                       \\ \hline
\multicolumn{1}{|c|}{Validation Set} &  {\includegraphics[scale=0.5]{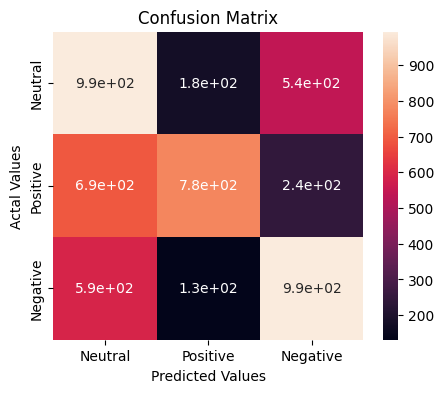}} & {\includegraphics[scale=0.5]{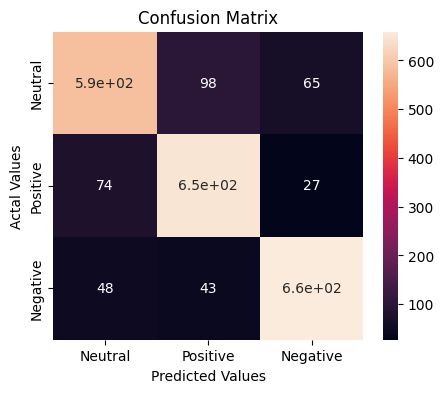}}  \\ \hline
Test Set                             &     {\includegraphics[scale=0.5]{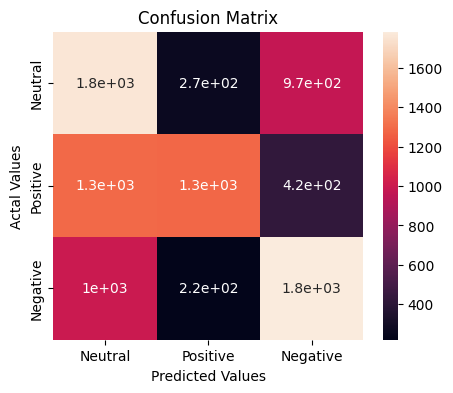}}       &         {\includegraphics[scale=0.5]{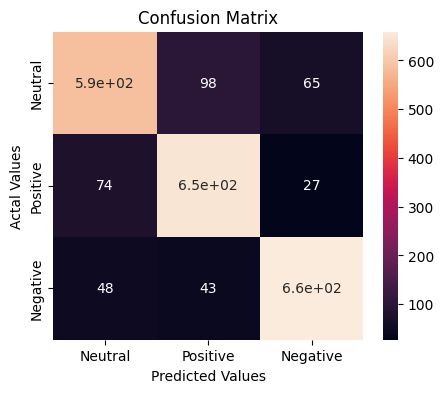}}                       \\ \hline
\end{tabular}

\label{tab:my-table}
\end{table*}
\begin{table*}
    
\caption{Confusion Matrices for Random Sequential Masking (MahaSent model)}
\centering
\begin{tabular}{|l|l|l|}

\hline
                                     & GoEmotions & MahaSent                       \\ \hline
\multicolumn{1}{|c|}{Validation Set} &  {\includegraphics[scale=0.5]{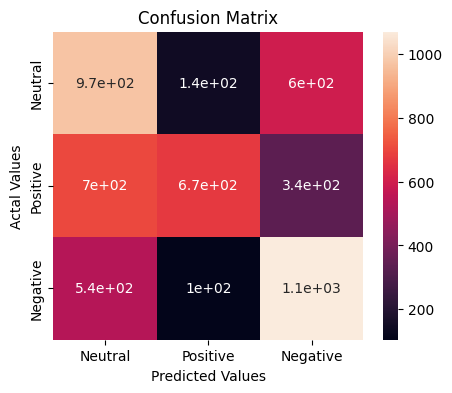}} & {\includegraphics[scale=0.5]{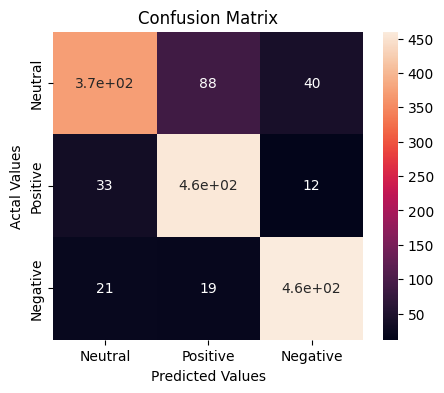}}  \\ \hline
Test Set                             &     {\includegraphics[scale=0.5]{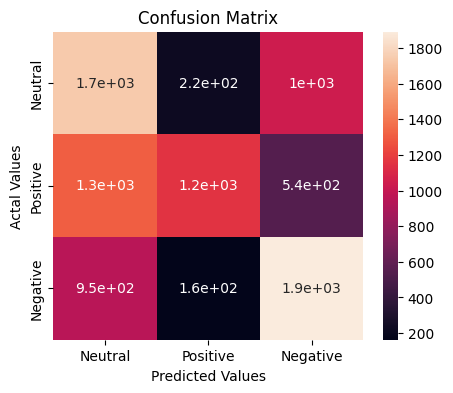}}     &         {\includegraphics[scale=0.5]{mahasent_sequential_m_val_cm.png}}                       \\ \hline
\end{tabular}

\label{tab:my-table}
\end{table*}
\section{Results}

The results of the models trained using data augmentation techniques are summarized in Table 1. These techniques aimed to enhance the performance of the models and improve their accuracy scores. Among the different approaches, the Named Entity Masking Using BERT(Sequential) technique exhibited the highest accuracy score of \textbf{0.5472} when the models were fine-tuned on the MahaSent dataset. This represented a significant improvement compared to the baseline models, particularly for the GoEmotions test set.

Interestingly, when the models were fine-tuned on the GoEmotions dataset, they achieved higher scores for the MahaSent test set, with an accuracy score of \textbf{0.7362}. This indicates that the models benefited from the knowledge transfer across domains, where the expertise gained from one dataset helped improve the accuracy on another.

The Random Masking Using BERT(Sequential) approach, which was trained on the MahaSent dataset, demonstrated a higher accuracy score of \textbf{0.8430} on the MahaSent test set. This represented an improvement from the baseline score of \textbf{0.8367}, indicating that the data augmentation technique successfully enhanced the model's ability to classify sentiment on this specific dataset.

For the model that was fine-tuned on GoEmotions, it achieved the highest accuracy of \textbf{0.6320}, indicating its effectiveness in accurately classifying sentiment within the context of the GoEmotions dataset.

These insights highlight the impact of data augmentation techniques on improving the performance of sentiment classification models. The results indicate that employing specific approaches such as Named Entity Masking Using BERT(Sequential) or Random Masking Using BERT(Sequential) can lead to notable improvements in accuracy scores, particularly when fine-tuning the models on specific datasets. Additionally, the findings suggest that knowledge transfer between domains can also contribute to enhanced accuracy when evaluating on different datasets.

The scores for the GPT approach 1 and approach 2 are presented in Tables 2 and 3, respectively. In particular, when conducting an in-domain analysis, the GPT approach 1, known as Label Generation, achieved the highest score of \textbf{0.8435} on the MahaSent test set. This result indicates that the GPT approach 1 outperformed all other models considered in this analysis. This finding highlights the effectiveness of the Label Generation approach in accurately generating sentiment labels for the MahaSent dataset. The GPT model, trained specifically for label generation, demonstrated its capability to understand and capture the sentiment expressed in the input sentences. This high score suggests that the GPT approach 1 can be a reliable and robust solution for sentiment analysis within the domain of the MahaSent dataset.   

Tables 4 to 7 contain the confusion matrices for all the approaches discussed.

\section{Conclusion and Future Work}

Our paper proposes a novel approach with respect to data augmentation for sentiment analysis in low-resource languages. Given the lack of available datasets in Marathi, we aim to leverage these techniques to address this problem. Through these data augmentation methodologies, we aim to augment existing datasets so as to enhance the quantity of labeled data for the sentiment analysis task.

\begin{itemize}
  \item The objective of our research was to improve the performance of Marathi sentiment analysis models using the proposed data augmentation techniques.
  \item We were able to not only enhance in-domain performance (accuracy on the same domain as the training data) but cross-domain performance (accuracy on different domain data) as well.
  \item By exploring both in-domain and cross-domain applications, we seek to demonstrate the effectiveness of our proposed data augmentation methods in overcoming the limitations imposed by the lack of datasets in Marathi sentiment analysis.
  \item This paper proposes data augmentation methodologies as a valuable contribution to the field, facilitating more accurate and comprehensive sentiment analysis in Marathi, along with a thorough analysis of the augmentation methods proposed.
\end{itemize}

The scope of the research focuses specifically on addressing the lack of datasets in sentiment analysis for Marathi only. The primary objective of the research work is to apply data augmentation methods to existing datasets and perform in-domain and cross-domain analysis. However, the paper’s scope is limited to bridging the gap by addressing the lack of datasets alone and does not encompass other areas such as the scarcity of preprocessing methodologies and other challenges in sentiment analysis for Marathi. \\

\par Furthermore, the focus was exclusively on research related to sentiment analysis in the Marathi language, and it does not extend to other languages. However, these techniques can be applied to other languages for further analysis of performance improvement and other sentiment analysis related research work.
  
\section*{Acknowledgments}
  This work was done under the L3Cube Pune mentorship
program. We would like to express our gratitude towards
our mentors at L3Cube for their continuous support and
encouragement.

\bibliographystyle{IEEEtran}
%\setlength{\bibsep}{3pt}
%\makeatletter
%\renewcommand\@biblabel[1]{[#1]}
%\makeatother
\bibliography{main}

\end{document}